\documentclass [12pt]{article} 

\usepackage{amsmath,amssymb,amsthm}
\usepackage{graphicx}
\usepackage{setspace}
\usepackage{indentfirst} 
\usepackage{float}
\usepackage{cite}
\usepackage{booktabs}

\usepackage{caption}
\graphicspath{ {images/} }

\theoremstyle{plain}

\theoremstyle{definition}

\begin{document}
\title{How facial features convey alertness in stationary environments}
\author{Janelle Domantay}
\date{\today}


\vspace{1.5in}

\begin{center}
    HOW FACIAL FEATURES CONVEY
ATTENTION IN STATIONARY ENVIRONMENTS
 \\
    By \\
    Janelle Domantay\\
    \vspace{1.5in}
    Honors Thesis submitted in partial fulfillment\\
    for the designation of Research and Creative Honors\\
    Department of Computer Science\\
    Dr. Brendan Morris\\
    Dr. William Doyle\\
    Dr. Jorge Fonseca \\
    Howard R. Hughes College of Engineering\\
    University of Nevada, Las Vegas\\
    December, 2021
\end{center}

\thispagestyle{empty}

\newpage

\begin{center}
    \Large
    
    \textbf{Acknowledgements}
\end{center}

I have a tremendous amount of gratitude for Dr. Brendan Morris as my primary faculty advisor and the first person to introduce me to research. Thank you for your support, advice and for all the opportunities you've provided me over the years; I can't thank you enough for the impact you've made on my development as a researcher.

I'd also like to thank my other committee members for their time and for providing feedback to my work in its earlier stages: Thank you Dr. Jorge Fonseca for providing additional insight and encouragement when I've needed it the most. Thank you Dr. William Doyle for providing a unique perspective and helping me navigate the Honors thesis process.

In addition, I'd like to thank Dr. Lisa Menegatos for the assistance she provided me during the course as well as the UNLV Honors College as a whole.



\newpage

\newpage

\begin{abstract}
Awareness detection technologies have been gaining traction in a variety of enterprises; most often used for driver fatigue detection, recent research has shifted towards using computer vision technologies to analyze user attention in environments such as online classrooms. This paper aims to extend previous research on distraction detection by analyzing which visual features contribute most to predicting awareness and fatigue. We utilized the open source facial analysis toolkit OpenFace in order to analyze visual data of subjects at varying levels of attentiveness. Then, using a Support-Vector Machine (SVM) we created several prediction models for user attention and identified Histogram of Oriented Gradients (HOG) and Action Units to be the greatest predictors of the features we tested. We also compared the performance of this SVM to deep learning approaches that utilize Convolutional and/or Recurrent neural networks (CNN's and CRNN's). Interestingly, CRNN's did not appear to perform significantly better than their CNN counterparts. While deep learning methods achieved greater prediction accuracy, SVMs utilized less resources and, using certain parameters, were able to approach the performance of deep learning methods. 
\end{abstract}

\newpage

\section{Introduction}

Facial analysis has been a steadily growing field. In particular, the reduction of auto-related accidents has become a major focus  for research related to fatigue and distraction detection. However, given the increasing shift towards computer-based work and learning enterprises, it has become more relevant to consider the efficacy and effects of prolonged attention in stationary environments. 

Through the use of visual information such as facial expression and eye closure, we can predict the attentiveness and fatigue of a user in a stationary environment which can be used to assess how to improve productivity and worker satisfaction. Utilizing the open source facial analysis toolkit, OpenFace \cite{openFacegit}, we analyzed video data sets \cite{dataset} of lab participants in order to gauge the effectiveness of different visual cues in predicting a user’s alertness. 

For the purposes of this study, we used awareness assessment technologies in order to study drowsiness of users. We define awareness based on a drowsiness scale constructed by \cite{dataset}. To simulate ground truth we define high alertness to be when a user self reports an absence of fatigue and is completely conscious\cite{dataset} while drowsiness is when a user self-reports requiring effort to stay awake.

The main factors we considered to classify a worker as alert, or fatigued include eyelid visibility \cite{gestureAndPerclos, gazeAndPerclos, perclos}, facial expression \cite{studentEmotion, deepLearn, FACS}, and Histogram of Oriented Gradients (HOG) \cite{hogCite}. Eyelid visibility will be used as a proxy for drowsiness and will be measured using an awareness assessment technology called PERCLOS\cite{perclos} which will be used to register a user’s eye and detect the changes in eyelid coverage. PERCLOS refers to the proportion/ percentage of the time in a minute that a subject’s eye is more than eighty percent closed. Facial expressions can also be used to determine fatigue and distraction, as previous research has indicated that non neutral facial expressions suggest that when a subject is dwelling on an emotion for longer than a given threshold \cite{studentAttention}, they may become inattentive to an assigned task.

\textbf{Contributions:} The contributions of this paper incldue building on previous studies on attention detection, by comparing the performance of different features in Support-Vector Machine (SVM) classifiers. Moreover, this paper compares the accuracy and processing time of Support-Vector Machine (SVM) classifiers against Convolutional Neural Networks (CNNs) and Convolutional Recurrent Neural Networks (CRNNs).

\section{Literature Review}

The majority of research pertaining to drowsiness detection has been conducted on human drivers in the interest of developing alerts that can prevent road accidents. However, given the recent shift to online learning, recent research has examined how attention detection can be employed in online learning environments.

\subsection{An Overview of Attention Metrics}
Methods for recognizing inattention can be categorized into contact, methods that utilize measurements from physical sensors; and non contact, methods that utilize visual aspects that can be recognized via webcam \cite{survey}. Recent research attempts also utilize data collected from both mediums, such as eye tracking data in conjunction with electroencephalogram signals \cite{survey}. However, since contact metrics can be more costly, inconvenient, and at times impossible, this paper seeks to develop models that enhance contact free metrics.

 Many studies have been conducted on measuring the attention or drowsiness of drivers through contact free methods \cite{survey, gestureAndPerclos, gazeAndPerclos, perclos, deepLearn, perclosPast}, particularly in relation to driver fatigue. Attentiveness and fatigue have also been manually measured through the use of Continuous Tracking Tasks (CTT), that monitor a user’s alertness based on their performance at a given task, and Psychomotoric Vigilance Tests (PVT), which monitors user reaction time by prompting users to interact with a flashing panel at random intervals \cite{perclos}. 

\subsection{Mapping Facial and Physical Motion}
Alertness measures can be inferred from Gjoreski et al.’s \cite{deepLearn} assessment of driver distraction. Specifically, they categorize distraction as either cognitive, emotional, sensorimotor, or mixed. While this study primarily focused on characteristics related to the face, it is noted that full body motion such as yawning, posture, and hand position were all cited to contribute to attentiveness. Similarly, Happy et al. \cite{studentEmotion} identified similar movements to be indicative of boredom or frustration when observed in students participating in a virtual classroom. While e-learning users are closer to our proposed demographic, the similarities in methodology and findings suggests that studies performed on driver inattentiveness is highly applicable to stationary environments. In regards to monitoring emotional distraction, Revadekar et al. \cite{studentAttention} performed a relevant study on gauging student attention in e-learning environments by measuring facial expression in addition to posture, lean, and head movement. Emotional inattentiveness was signified by a detected emotion enduring for longer than a given threshold which suggested that a user’s attention was directed away from the assigned task. The aforementioned study uses a threshold of ten minutes.

Popular metrics for measuring facial expression also come from the use of the Facial Action Coding System (FACS), and the Histogram of Oriented Gradients (HOG) \cite{hogCite, FACS}. FACS are a standardized method for categorizing facial muscle movement developed by Carl-Herman Hjortsjö. The current standard for FACS was developed by Paul Ekman and Wallace Friesen and is currently the industry standard \cite{FACS}. In facial analysis, HOG features are extracted by encoding facial feature components into a single vector which can then be fed into an SVM \cite{hogCite}. 

\subsection{Eye Mapping}
In regards to eye mapping, Madsen et al. \cite{studentSync} monitored eye movement and pupil size from subjects who were viewing visual stimuli with and without a distraction task. Madsen et al. found that subjects that were attentive to their task tended towards synchronized eye movement, while subjects that were distracted had erratic movement. In addition, subjects with synchronized eye movement also tended to recall more information from the visual stimuli when tested on it in comparison to their distracted counterparts.

One of the most popular metrics for measuring alertness is PERCLOS, an alertness measure founded in the 1990s that predicts fatigue in relation to the percent closure of a user’s eye. Touted as the most promising and “first-ever” real-time drowsiness detection sensor \cite{perclos}, PERCLOS has become a popular tool in subsequent studies. PERCLOS consists of three drowsiness metrics: P70, the proportion of time the eyes were closed at least 70 percent; P80, the proportion of time the eyes were closed at least 80 percent; EYEMEAS (EM), the mean square percentage of the eyelid closure rating \cite{perclos}. Of these three, P80 was cited to correlate best to driver fatigue \cite{gestureAndPerclos}. 

While numerous studies continue to be performed using PERCLOS, a recent study by Trutschel et al. \cite{perclosPast} cites that alternative measure technologies such as Eye-Tracking Signal and contact measurements via EGG and EOG recordings yield better performance at identifying lapses in attention that occur over a shorter period of time. Since the results of this study are meant to analyze a general decrease in attentiveness as opposed to momentary lapses, we have opted to use PERCLOS for analysis purposes. 

\subsection{Facial Analysis Toolkits}
Extensive research has also been conducted on models for accurately mapping facial movement in realtime. Toolkits such as Affectiva and OpenFace have been used both for mapping head gesture and recognizing facial expressions. Affectiva, while adept in emotion recognition, has primarily been used in commercial settings \cite{affectiva}. 

Alternatively, OpenFace has become popular as an open source toolkit for mapping facial expression, gaze direction, and head pose with real-time performance. Utilizing Conditional Local Neural Fields (CLNF), OpenFace employs a Point Distribution Model to outline the shapes of various facial landmarks in order to recognize and map their location and motion. Additional points are fitted around the eyes, lips, and eyebrows. OpenFace also uses a three layer Convolutional Neural Network (CNN) that was trained using datasets of faces from various angles. By training on unconventional face angles, the CNN allows OpenFace to account for landmark detection error. The CLNF framework is also used to extract head pose and eye gaze. Head pose is mapped in a similar way to the aforementioned facial landmark detection, and eye gaze is mapped by using CLNF to locate the eye and pupil to compute a gaze vector \cite{openFace}. Facial expression is determined by the use of select Action Units, listed in Table \ref{tab:FACS}, which correspond to their respective FACS\cite{FACS} facial movements. OpenFace detects the presence of all listed Action Units on a binary scale (0 or 1), and detects the intensity of all Action Units on a 1 to 5 scale with the exception of AU28.

\begin{table}[t]
\centering
\caption{Action Units detected by OpenFace}
\label{tab:FACS}
\begin{tabular}{cc}
\toprule
\textbf{Action Unit} & \textbf{Facial Feature}  \\ 
\midrule
AU01 & Inner brow raiser \\
AU02 & Outer brow raiser \\
AU04 & Brow lowerer \\
AU05 & Upper lid raiser \\
AU06 & Cheek raiser \\
AU07 & Lid tightener \\
AU09 & Nose wrinkler \\
AU10 & Upper lip raiser \\
AU12 & Lip corner puller \\
AU14 & Dimpler \\
AU15 & Lip corner depressor \\
AU17 & Chin raiser \\
AU20 & Lip stretcher \\
AU23 & Lip tightener \\ 
AU25 & Lips part \\
AU26 & Jaw drop \\
AU28* & Lip Suck \\
AU45 & Blink \\
\bottomrule
\end{tabular}\\
\footnotesize{*Can only be detected by presence (0 or 1)}
\end{table}

\subsection{Machine Learning Methods}
In addition to available toolkits, researchers have also yielded results by classifying facial features in relation to the centroid of the human face. By locating facial landmarks, boosting algorithms such as Adaboost can be used to reliably track head gesture in real time \cite{gestureAndPerclos}. After establishing the face’s centroid, deviations in head position can be tracked by calculating euclidean distance between the average centroid and the current centroid. When a subject’s head deviates from the average for extended periods of time we expect this to be indicative of distraction.

Methods used to extract visual features and build  training models generally stem from deep learning approaches \cite{survey} which utilize neural networks containing multiple hidden layers. Well suited for image classification, deep learning models such as Convolution Neural Networks (CNN), Recurrent Neural Networks (RNN), and Deep Belief Networks (DBN) are among the most common methods used for fatigue and distraction detection in human driver studies \cite{survey}. Deep learning  or deep neural network models refer to Artificial Neural Networks (ANN), which consist of feeding input into a series of hidden layers in order to produce an output \cite{rnn, introML}. CNNs are distinguished by their convolutional layer, which performs matrix operations on inputted data in order to reduce image sizes \cite{cnn}. RNNs are distinguished by their ability to take previous outputs into consideration via feedback loops \cite{rnn}. For video classification, CNN's and RNN's are commonly used in conjunction to form a Convolutional Recurrent Neural Network, or CRNN, wherein the CNN performs feature extraction on individual video frames, and then is reshaped in order to feed into the RNN. This approach combines the CNN's adeptness at image processing with the RNN's capacity for memory. Furthermore, Support-Vector Machines (SVM), are often used as classifiers for extracted features \cite{gazeAndPerclos, introML}. An SVM, or kernel machine refers to a model of classification that is derived from calculating the discriminant of support vectors \cite{introML}. An SVM's kernel determines what kinds of mathematical functions are employed to transform the data.

\subsection{Additional Neural Networks}
Common neural networks utilized in computer vision, include Mobile Networks (MobileNets) \cite{mobilenets}, Residual Networks (Resnets)\cite{resnet}, Inception networks (Inception)\cite{inception}, and Dense Convolutional Networks (DenseNet)\cite{densenet}. These networks can be implemented into an existing CNN, RNN, and CRNN. 
MobileNets refer to a class of neural networks designed with a focus on optimization and efficiency. They are composed of factorized convolutions which are meant to reduce the model's size and computation \cite{mobilenets}. ResNet refers to a residual neural network which utilizes a deep residual learning framework composed of residual blocks that use "shortcut connections" to add the outputs of previous layers directly to the outputs of stacked layers. This framework operates on the  The Inception\cite{inception} neural network is a deep convolutional neural network characterized by the use of convolutions of varying sizes (1 x 1, 3 x 3, and 5 x 5 in the presented paper) which are then layered over one another. InceptionV3\cite{inceptionv3} is an upgraded version of the initial model which incorporates factorized convolution sizes in place of larger convolutions and batch-normalization of auxiliary classifiers. The Dense Convolutional Network\cite{densenet}, or DenseNet architecture connects every layer of the neural network to every other layer.

\section{Research Questions}
Generally, we seek to analyze how visual information can be used to predict attentiveness and fatigue in stationary workers. Thus, our research question asks “Which visual features are more well suited for predicting either attentiveness or fatigue?” Since our research question centers around features, the ideal method for building a prediction model would be through an SVM due to the nature of its classification model, namely, its necessity for explicit feature extraction. Additionally, we are interested in comparing traditional machine learning methods that utilize SVMs to deep learning models such as CNNs and RNNs. Moreover, due to the nature of hidden layer neural networks, it is more difficult to identify which visual features contribute to a particular outcome. This provides much less explainability for CNN and RNN outcomes, whereas using an SVM is ideal for specifically identifying which features contribute to drowsiness prediction.

\section{Methodology}
Our goal is to compare the performance of three different approaches, SVM's, CNN's, and CRNN's, at predicting drowsiness given visual input.

\subsection{Classical Machine Learning}

\begin{figure}[h]
\centering
\caption{Using an open source facial analysis toolkit, we can extract visual features from imported video data which we will use to train our SVM classifiers.}
\label{fig:svm}
\includegraphics[scale = .25]{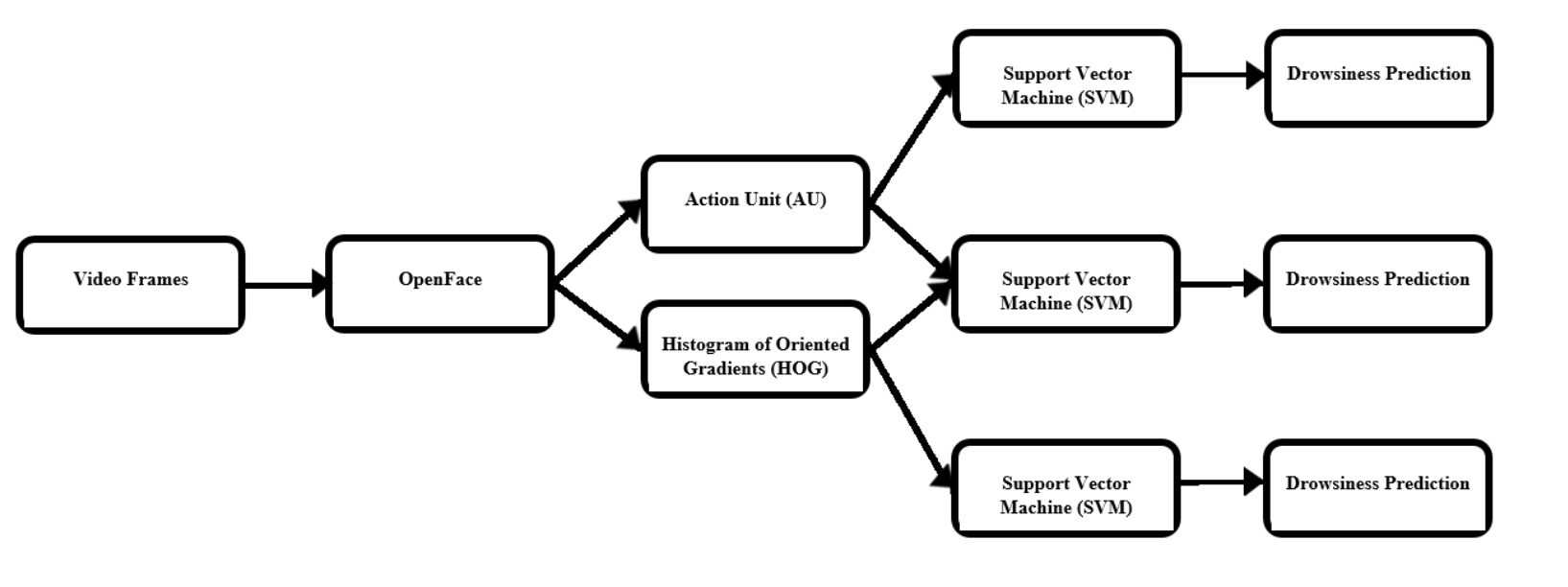}

\end{figure}

Figure \ref{fig:svm} provides an overview of our classical approach to drowsiness detection. We input video frames through OpenFace in order to extract feature vectors. These feature vectors are then used to classified into a drowsiness level using an SVM. 

Utilizing OpenFace, we were able to track facial landmark movement, gaze vector, and facial Action Units  such that landmark movement and gaze vectors are used to convey head gesture and gaze information respectively. The facial Action Units, account for facial muscle movements and eye closure. Data was collected from every frame of our constructed subset. To approximate PERCLOS, we utilize AU45\_r. While the FACS specifies that AU45 to refer to blinks, OpenFace captures the intensity of certain action units on a scale of 0 to 5.

Through a pilot study\footnote{Pilot Study and additional information recorded in Appendix.}, we determined that the visual features that best predicted attention included HOG and AU's. Therefore, the remainder of our studies utilized HOG and AU's over a subset of video frames from the UTA data set. The AU features we were able to extract using OpenFace are listed on Table \ref{tab:FACS}

Using the data set described in the previous section, we processed each of the video frames utilizing OpenFace and used the exported features to train a myriad of SVM's. The results are described in Table \ref{tab:svm_subset}. We also investigated how video segments affected SVM classification performance. Utilizing a subset of 28 frame video sequences, we extracted AU and HOG features that were averaged across the entire video. We then inputted these features into an SVM, the results of which are described in Table \ref{tab:svm_video}

\subsection{Deep Learning}

In contrast to our classical machine learning approaches, deep learning is used for end-to-end training. Images are directly inputted into our model such that the feature extraction and classification occurs in a single neural network. In our work, we compare both image based approaches, utilizing CNNs, and video, utilizing CRNNs.

\subsubsection{Convolutional Neural Networks}
\begin{figure}[t]

\centering
\caption{When using CNN's, we can directly feed visual input into the neural network. The CNN will perform feature extraction and drowsiness classification internally.}
\label{fig:cnn}
\includegraphics[scale = .3]{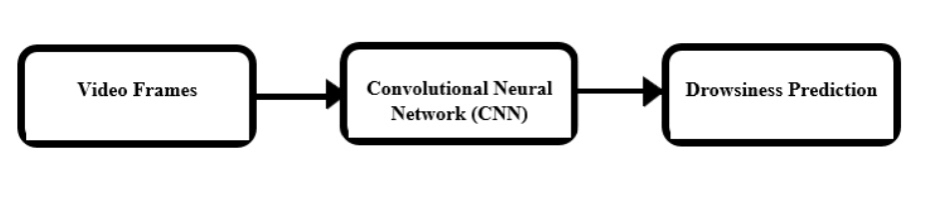}

\end{figure}

Figure \ref{fig:cnn} showcases our data pipeline for CNN's. Individual video frames are fed into the network, where feature extraction and classification is automated by the neural network.

To investigate the performance of neural networks on our data set, we began by using the imageAI \cite{ImageAI} toolkit, which utilizes Tensorflow and Keras CNN models in order to make predictions based on input images. The models tested included MobileNetV2\cite{mobilenets}, ResNet50\cite{resnet}, InceptionV3\cite{inceptionv3}, and DenseNet121\cite{densenet}, all of which were trained from scratch using the UTA subset over 100 epochs. Though not state-of-the-art, these networks are typical backbones and provide a good baseline for expected performance.  The results of these models are depicted in Table \ref{tab:cnn_scratch}.  

While training from scratch can produce highly specific models for drowsiness detection, it generally requires large volumes of data and significant computational resources to train.  When there is a data deficiency, transfer learning help speed up the process by starting from a well-trained model for general classification problems (e.g. general object recognition).  The network can then be modified for the particular task (drowsiness detection).  In this way, strong feature extraction comes from the large general recognition training while more high-level and problem specific features can be learned quickly.

To compare the performance of transfer learning, we utilized code from the Transfer-Learning-Suite \cite{TransferToolkit} created by George Seif, which utilizes Keras to perform training with pretrained models. Our pretrained models were trained using the ImageNet \cite{imagenet} and were provided by Keras. Using transfer learning, we can work from a model that has some experience in recognizing and classifying images rather than training a model to analyze an image from scratch. This may be lead to a quicker convergence rate or higher validation accuracy. We utilized the same neural networks as the CNNs trained from scratch for consistency. The results of our transfer learning models are recorded in Table \ref{tab:cnn_transfer}.

\subsubsection{Convolutional Recurrent Neural Networks}

\begin{figure}[t]
\centering
\caption{The CNN performs feature extraction on individual video frames. These images are then inputted into a recurrent neural network as 28 frame sequences. These sequences are analyzed holistically to determine a drowsiness classification.}
\label{fig:crnn}
\includegraphics[scale = .3]{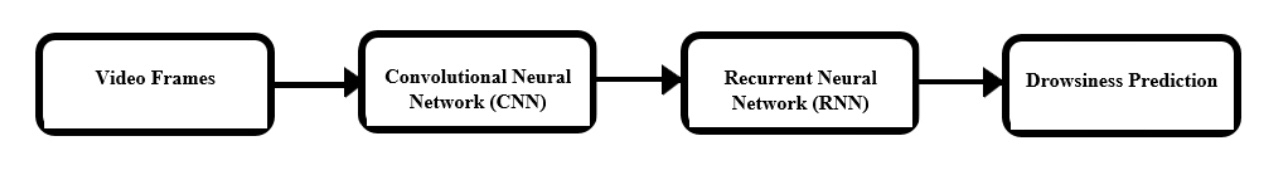}

\end{figure}

Wanting our final comparison point to be RNN's, we utilized the video-classification repository provided by Huan-Hsin Tseng \cite{videoClassifier} in order to run our data on CRNN models. As with our previous models, an overview of our model is depicted in Figure \ref{fig:crnn}. The CRNN is composed of both a CNN and RNN. We utilize the CNN portion for feature extraction on individual video frames. The feature vectors outputted from the CNN are then fed into the RNN which utilizes their internal memory to analyze multiple feature vectors for classification. The RNN's capacity for analyzing multiple video frames at a time allow us to explicitly model spatio-temporal features, which may be more beneficial for predicting drowsiness as opposed to the single image classifications that our previous CNN model performs. The CRNN models would then evaluate each video segment of our subset individually based on the input of 28 consecutive frames. The models utilized included a CRNN trained from scratch and a CRNN model that utilized a ResNet-152 neural network that was pretrained on the ILSVRC-2012-CLS ImageNet dataset \cite{imagenet}. 

For comparison, we also recorded results from a 3D CNN. Rather than relying on an RNN to account for temporal variation of a video sequence, a 3D CNN analyzes a video sequence as a 3D tensor of images, where the depth of the tensor is equal to the number of frames of the video sequence.

All of these results can be found on Table \ref{tab:rnn}.

\section{Experimental Evaluation}

\begin{table}[h]
\small
\centering
\caption{A drowsiness scale designed by \cite{dataset} for the purposes of their study.}
\label{tab:drowsyScale}
\begin{tabular}{lccc}
\toprule
\textbf{Scale Value} & \textbf{Qualitative Description} & \textbf{Classification} \\ 
\midrule
   1             & Extremely Alert  & Alert \\ 
   2             & Very Alert       & Alert \\
   3             & Alert & Alert                      \\
    4         & Rather Alert                              \\ 
   5             & Neither Alert or Sleepy                           \\ 
   6             & Some signs of sleepiness & Low Vigilance                      \\ 
    7             & Sleepy, but no difficulty remaining awake & Low Vigilance                              \\ 
   8             & Sleepy, some effort to keep alert & Drowsy                           \\ 
   9             & Extremely Sleepy, fighting sleep & Drowsy                      \\ 

\bottomrule
\end{tabular}
\end{table}

Our models were all trained using the UTA Real-Life Drowsiness Dataset. Performance was characterized by Test/Validation accuracy, the proportion of correctly classified data instances from the test set, as well as processing time. Additional metrics for performance included precision and recall for our SVMs, and loss and epochs for our deep learning methods. 

\subsection{UTA-RLDD Dataset}

\begin{figure}[t]

\centering
\caption{A sample of video frames from the UTA-RLDD. Image taken from \cite{dataset}. Categorized as alert (first, row), low vigilance (second row) and drowsy (third row).}
\includegraphics[scale = .4]{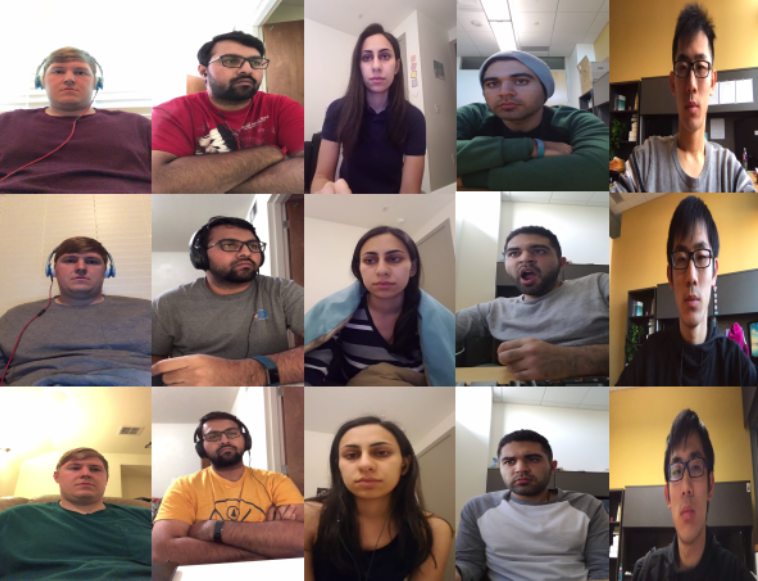}
\label{fig:uta}
\end{figure}

We utilized the open source video data set developed by R. Ghoddoosian et al. \cite{dataset}, referred to as the University of Texas at Arlington Real-Life Drowsiness Dataset (UTA-RLDD). Featuring sixty participants recording themselves at self reported levels of alertness or drowsiness, UTA-RLDD serves as an  ideal resource for testing our models. Each video featured approximately ten minutes of a participant in a self-reported state of alertness, classified as either "Alert," "Low Vigilance", or "Drowsy." Participants classified themselves into one of these three classes according to a nine point scale wherein 1, 2, and 3 mapped to "Alert", 6 and 7 mapped to "Low Vigilance," and 8 and 9 represented "Drowsy" as shown in Table \ref{tab:drowsyScale}. A sample of video frames from this data set are depicted in Figure \ref{fig:uta}. Most videos consist of user videos taken from the front. Based on this data we built a variety of models using both deep learning and classical machine learning methods. 

In the interest of time and convenience, we opted out of using the entirety of the data set. Moreover, during pilot studies we conducted utilizing the entire data set, we did not find a significant difference between the performances of models trained on the entire dataset, and those trained on a subset. Consequently, we constructed a subset of 3000 frames, with 1000 frames selected from each class, which we used to train and test each of our models. To collect these frames, we first extracted video frames in the form of jpeg files from each video of the data set. To control for each video's length, and to discard data that might be associated with the beginning and end of a video, we only selected video frames between frame 1440 (roughly one minute) to 14000 (roughly nine minutes). For each class, we then generated 1000 timestamps that were equally dispersed throughout this eight minute period, and randomly selected images from our pool of video frames. 

\begin{table}[h]
\small
\centering
\caption{Data set Specifications}
\label{tab:dataset}
\begin{tabular}{lcccc}
\toprule
\textbf{Data set} & \textbf{Total Images} & \textbf{Train} & \textbf{Test} \\ 
\midrule
   UTA-RLDD             & 3076270            &   - & -                     \\ 
   Image Subset             & 3000              &  2142 & 858                  \\ 
   Video Subset             & 84000            & 59976 &  24024           \\ 

\bottomrule
\end{tabular}
\end{table}

In order to account for the video segments we would use during our studies with CRNN's, we also created a subset of video segments that were composed of each frame from the image subset along with the twenty-seven frames that preceded it. Our training data is composed of video frames that were selected before frame 10000, roughly the seven minute mark, while our testing data is composed of frames collected after this point. These specifications of these data sets are recorded in Table \ref{tab:dataset}. It should be noted that based on our experimental design, our models do not perform predictions on users that it hasn't already encountered during training \footnote{Additional research was conducted that tested our models on individuals who were not present in the data set. The results of such are insubstantial, but are present in the appendix and may serve as a future research direction.}.

\subsection{Evaluation Metrics}
All models were evaluated based on the proportion of images from the test set that were correctly classified as either alert, low vigilance, or drowsy. 

In addition to analyzing the accuracy of each of our models, we also analyzed the processing time of each model. This was obtained by timing the number of seconds it took for each model to classify a single image, or in the case of the CRNN models, a video sequence. All models were timed on the same machine, but it should be noted that the neural networks were run on a graphical processing unit (GPU), while the SVM models only ran on a machine's central processing unit (CPU). It should be noted that the processes for SVMs were timed using milliseconds, while the deep learning processes were timed using seconds.

Classical learning methods were gauged via their precision and recall for each class. Precision and recall were calculated by the weighted average of the individual classes. 

We refer to the classes alert, low vigilance, and drowsy as the by the values 0, 1, and 2 respectively in the following formulas. 
Weighted precision is calculated by
\[ precision = \frac{(w_0 * \frac{tp_0}{tp_0 + fp_0}) + (w_1 * \frac{tp_1}{tp_1 + fp_1}) + (w_2 * \frac{tp_2}{tp_2 + fp_2})}{w_0 + w_1 + w_2}\]
such that for $0 <= i <= 2$, $w_i$ refers to the number of instances in class $i$, $tp_i$ refers to the number of instances that the model correctly classified into class $i$, and $fp_i$ refers to the number of instances that were incorrectly classified into class $i$. Similarly, weighted recall is calculated by
\[ recall = \frac{(w_0 * \frac{tp_0}{tp_0 + fn_0}) + (w_1 * \frac{tp_1}{tp_1 + fn_1}) + (w_2 * \frac{tp_2}{tp_2 + fn_2})}{w_0 + w_1 + w_2}\]
where $fn_i$ refers to the number of instances that belong to class $i$ but were not correctly labelled by the classifier. Colloquially, we refer to the values $tp$, $fp$, and $fn$ as true positives, false positives, and false negatives respectively. 

The CNN and CRNN models were evaluated based on their training accuracy, testing accuracy, and loss. Training accuracy refers to the accuracy obtained from the model classifying the data set it was trained over, while test accuracy refers to the model's ability to classify the test data set. A loss function is a function used to determine a model's error in classification by comparing the model's expected output to its actual output. It generalizes this comparison by outputting loss, which refers to a score given to a model that serves as a summation of the model's error in incorrectly classifying data by comparing the expected output to the model's actual. Thus a model's objective is to minimize its loss score. Overtime, as the model improves at classifying the data set, the loss value should drop. A low loss score generally correlates to a higher performing model. 

\subsection{Implementation Details}
All models were run on Ubuntu 18.04.4 using Python 3.7.6, PyTorch 1.8.1 and Keras 2.4.3. The classification models were run entirely on an Intel(R) Xeon(R) Gold 5218 CPU, while the deep learning models utilized a Quadro RTX 6000 GPU. 
The CNN models trained from scratch were trained at an initial learning rate of .001, while the models trained via transfer learning began with a learning rate of 0.00001. Both utilized a categorical cross-entropy loss function. 

The CRNN model trained from scratch used an initial learning rate of 0.0001, while the transfer learning model used a learning rate of 0.001. Both CRNN models and the 3DCNN model utilized a cross entropy loss function. 

CNN and CRNN models that were trained from scratch were run across 100 epochs, which was an upper limit for convergence across all networks. Pretrained CNN models converged within 20 epochs and were only run for that amount. In contrast, the ResNetCRNN model converged at a much slower rate and was also run at 100 epochs to converge. 

The 3D CNN was run for 15 epochs and converged the quickest out of all deep-learning models. 

\subsection{Classical Results}
\begin{table}[h]
\small
\centering
\caption{Support-Vector Machine Results}
\label{tab:svm_subset}
\begin{tabular}{lcccccc}
\toprule
\textbf{Attributes} & \textbf{Kernel} & \textbf{Precision} & \textbf{Recall} & \textbf{Test Acc.} & \textbf{Time(ms)} \\ 
\midrule
AU   & RFC*           & 0.4685              & 0.4732        & 0.4732 & 4.332E-3        \\   
  & Linear             & 0.4719              & 0.4685        & 0.4685  &  0.01254                \\ 
  & Polynomial             & 0.6762              & 0.6282        & 0.6282    & 0.01136               \\ 
  & Sigmoid             & 0.3747              & 0.3753        & 0.3753  & 0.02650      \\ 
   & Gaussian         & 0.7144              & 0.7145        & 0.7145    & 0.03168                \\  
   \midrule
HOG   & RFC            & 0.6291              &  0.6270       & 0.6270      &   0.1059                 \\   
  & Linear             & 0.9067              & 0.9068        & 0.9068   & 0.9203                \\ 
   & Polynomial             & 0.9551              & 0.9545        & 0.9545   &  1.386                  \\ 
   & Sigmoid             & 0.5076              & 0.4848        & 0.4848    & 1.761                  \\ 
  & Gaussian         & 0.9307              & 0.9301        & 0.9301     & 2.316             \\ 
  \midrule
HOG \& AU   & RFC           & 0.6034              & 0.6014        & 0.6014   & 0.1097        \\   
   & Linear             & 0.9103             & 0.9103       & 0.9103        &   1.011            \\ 
  & Polynomial             & 0.9563              & 0.9557        & 0.9557    & 1.313                  \\ 
 & Sigmoid             & 0.4660              & 0.4545        & 0.4545   &  1.663               \\ 
 & Gaussian         & 0.9271              & 0.9266       & 0.9266     & 2.338                \\
\bottomrule
\end{tabular}
\flushleft
\footnotesize{*Random Forest Classifier}
\end{table}

\begin{table}[h]
\small
\centering
\caption{Support-Vector Machine Results using Video}
\label{tab:svm_video}
\begin{tabular}{lcccccc}
\toprule
\textbf{Attributes} & \textbf{Kernel} & \textbf{Precision} & \textbf{Recall} & \textbf{Test Acc.} & \textbf{Time(ms)} \\ 
\midrule
AU   & RFC           & 0.4095             & 0.4103        & 0.4103 & 4.822E-3        \\   
  & Linear             & 0.3520              & 0.3504        & 0.3520  &  0.01348                  \\ 
  & Polynomial             & 0.4563              & 0.4219        & 0.4219    &   0.01106             \\ 
  & Sigmoid             & 0.3716              & 0.3660        & 0.3660  &   0.02695      \\ 
   & Gaussian         & 0.4601              & 0.4580        & 0.4580    &    0.03235             \\  
   \midrule
HOG   & RFC            & 0.6104              &  0.6010       & 0.6011      & 0.1057               \\   
  & Linear             & 0.5291              & 0.5245        & 0.5245   &    1.651            \\ 
   & Polynomial             & 0.9431              & 0.9429        & 0.9429   & 1.312                   \\ 
   & Sigmoid             & 0.9282              & 0.9277        & 0.9277    & 1.688                  \\ 
  & Gaussian         & 0.4815             & 0.4464        & 0.4464     &    2.326        \\ 
  \midrule
HOG \& AU   & RFC           & 0.6770              & 0.6760        & 0.6760   & 0.1040            \\   
   & Linear             & 0.9488             & 0.9487       & 0.9487        &  1.663            \\ 
  & Polynomial              & 0.9604             & 0.9604       & 0.9604     &  1.264                   \\ 
 & Sigmoid             & 0.5045             & 0.4709        & 0.4709   &  1.671                  \\ 
 & Gaussian         & 0.9467              & 0.9464       & 0.9464     &   2.337              \\ 

\bottomrule
\end{tabular}
\end{table}
The classifiers' overall performance (Table \ref{tab:svm_subset}) was characterized by classifying a testing data set which was comprised of 20 percent of select data. In general, the attributes that best predicted drowsiness were HOG. The Polynomial and Gaussian kernel correlated to highest validation accuracy, while the Sigmoid correlated to the worst. The models which utilize HOG and AU attributes in conjunction with a Polynomial kernel most closely approach the accuracy attained by the deep learning methods. It should be noted, however, that the model that only utilizes HOG attributes attain nearly as high of a validation accuracy, and the high performance of the "HOG \& AU" model may be dependant on the presence of HOG features rather than the conjunction of HOG with AU's. 

The SVM models that utilized video segments, as depicted in Table \ref{tab:svm_video}, did not perform significantly better than its counterpart, and actually performs worse under certain parameters. One notable difference, however, is that it achieves an accuracy of 96.04\% when trained on both HOG and AU data over a polynomial kernel. In contrast to its predecessor, the SVM models that analyze multiple images at a time appears to benefit from having multiple attributes to train on rather than just HOG.

\subsection{Deep Learning Results}
The performance for CNN variants are summarized in Tables \ref{tab:cnn_scratch} and \ref{tab:cnn_transfer}.  

\subsubsection{Convolutional Neural Networks}

\begin{table}[h]
\small
\centering
\caption{CNN Training From Scratch}
\label{tab:cnn_scratch}
\begin{tabular}{lcccccc}
\toprule
\textbf{CNN} & \textbf{Epochs} & \textbf{Acc.} & \textbf{Loss} & \textbf{Val.} & \textbf{V. Loss} & \textbf{Time(s)} \\ 
\midrule
MobileNetV2    & 100             & 0.9993              & 0.0041        & 0.6695                       & 1.0328 & 1.955\\   
ResNet50     & 100             & 0.9980              & 0.0064      & 0.9880                       & 0.0355 & 2.131\\
DenseNet121  & 100             & 1.00              & 0.0057        & 0.9896                       & 0.0453 & 2.866\\ 
InceptionV3  & 100            & 0.9973              & 0.0055      & 0.9964                       & 0.0142 & 2.560\\
\bottomrule
\end{tabular}
\end{table}

\begin{table}[h]
\small
\centering
\caption{CNN Training Results using Transfer Learning}
\label{tab:cnn_transfer}
\begin{tabular}{lcccccc}
\toprule
\textbf{CNN} & \textbf{Epochs} & \textbf{Acc.} & \textbf{Loss} & \textbf{Val.} & \textbf{V. Loss} &  \textbf{Time(s)}\\ 
\midrule
MobileNetV2    & 20             & 0.9983              & 0.0110        & 0.9916                       & 0.0308 & 2.007\\   
ResNet50     & 20             & 0.9985              & 0.0088      & 0.9928                       & 0.0192 & 2.025
\\
DenseNet121  & 20             & 0.9992              & 0.0134        & 0.9784                       & 0.0634 & 3.354\\ 
InceptionV3  & 20           & 0.9960             & 0.0279      & 0.9892                       & 0.0441 & 2.581\\
\bottomrule
\end{tabular}
\end{table}

In comparison to their transfer learning counterparts, which tended to converge in only 20 epochs, the models from scratch reached similar levels of accuracy but required many more iterations. In contrast, the transfer learning models, while converging more quickly, required a slightly longer processing time. Overall, the models with the highest validation accuracy were the InceptionV3 model when trained from scratch, and the ResNet50 model when trained via transfer learning. The lowest performing model was the MobileNetV2 when trained from scratch, having a 67 percent validation accuracy after convergence. Interestingly, this issue is resolved in its transfer learning counterpart. 

Generally speaking, the transfer learning models performed better overall in comparison to the models trained from scratch, but with the exception of MobileNetV2, both methods can be used to achieve approximately the same accuracy. 

\subsubsection{Recurrent Neural Networks}
The comparison of CRNN (standard convolutional architecture CRNN and a ResNet-based backbone ResNetCRNN) and 3DCNN are shown in Table \ref{tab:rnn}.  The performance for all three variants are are quite strong -- perfect training accuracy and very high validation accuracy.  The ResNetCRNN had the best performance as expected since it has the most backbone capacity.  However, ResNetCRNN was by far the slowest of all the classification methods.  The simple CRNN had strong performance with a fraction of the processing time, making it a strong candidate for real-time implementation.

\begin{table}[h]
\small
\centering
\caption{RNN Training Results}
\label{tab:rnn}
\begin{tabular}{lcccccc}
\toprule
\textbf{CRNN} & \textbf{Epochs} & \textbf{Acc.} & \textbf{Loss} & \textbf{Val.} & \textbf{Val. Loss} & \textbf{Time(s)*}\\ 
\midrule
CRNN    & 100             & 1.000              & 1.745E-4        & 0.9697                       & 0.2158 & 0.05449\\   
ResNetCRNN     & 100             & 1.000              & 7.227E-4      & 0.9848                       & 0.0992 & 14.36\\
3DCNN  & 15             & 1.000              & 0.0201        & 0.9545                       & 0.1609 & 0.2814\\ 
\bottomrule
\end{tabular}\\
\flushleft
\footnotesize{*per 28 frame sequence}
\end{table}

It should be noted that although the ResNetCRNN model utilizes transfer learning, it converged at a slower rate than the transfer learning methods used by the CNN's, it is for this reason that the ResNetCRNN was also allowed to iterate for 100 epochs. In contrast, the 3DCNN converged at a much quicker rate, so it was run for 15 epochs as recommended by \cite{videoClassifier}.

\subsection{Comparison}
While it is clear that the neural network models achieved overall greater accuracy, it should be noted that feature extraction for these models does not solely rely on a user's facial expression. In contrast, the SVM's, while having lower test accuracy, were still able to achieve relatively high results by relying on data extracted from an individual's face. 

Additionally, the classical ML approach with SVM has significantly lower computational cost at around 1-2 ms per image while the DL methods were on the order of seconds.  

Based on this experiment, the SVM approach would be more desirable than DL for real-time drowsiness detection.  However, further experiments should be performed with more images and more participants to verify generalizability of all the techniques.

\section{Conclusion}
Based on our observations, while neural networks generally performed better attention classification, it is possible for SVM models to approach the accuracy of the deep learning models at a much lower processing time. We also believe that the methodology used to assess the performance of these classical models may be re-purposed for future research. The CNN models appeared to have the highest performance and quickest convergence in comparison to the CRNN counterparts, this is particularly interesting as it implies that drowsiness can be determined based on snapshots of an individual over a period of time. The CRNN models, while converging at a slower rate, performed similarly to the CNNs, but required a higher processing time. The highest performing SVM utilized the averaged HOG and FACS attributes of video segment frames over a Polynomial kernel. 

Based on extraneous research described in the appendix, we hypothesize that the performance of these models is highly dependant on whether a model has seen an individual before. Therefore, future research would involve testing these models on a dataset composed of more individuals and testing these models on subjects not present in the training set. 
\newpage

\bibliographystyle{unsrt}
\bibliography{sources}

\newpage

\section{Appendix} 

\subsection{Pilot Study}

Our pilot study consisted of testing six videos out of the UTA dataset in order to assess how gaze information, eye landmarks, and head pose all contributed to attention prediction. Though an informal study, there was still a significant improvement in model accuracy when using Action Units and HOG over the other features. When used independently or in conjunction, eye landmarks, head pose, and gaze information all performed at or slightly below 33 percent. Moreover, any additional models that combined Action Units with any of the aforementioned features also had much lower accuracy in comparison to models that solely used Action Units.  

\subsection{Challenges and future research}

Although not included in our primary study, we did investigate the performance of the CNN and RNN neural networks on a alternate subset. This alternate subset was constructed as follows: 

In R. Ghoddoosian et al.'s \cite {dataset} data set, videos are divided into groups of twelve called Folds with the intention of users using four of these folds for training data, and the remaining fold for testing. We follow this convention throughout the remainder of our study, by subdividing our subsets into folds based on which fold each frame or video segment originates from. Upon testing, we discovered that each of our previously described deep learning method performed poorly, averaging a 33 percent accuracy rate. We hypothesize that this is due to the fact that our training data no longer contains information on all individuals from the sample set and thus is unable to make accurate predictions when given a new person's video data. 

\end{document}